# ADVERSARIAL ATTACK ON YOLOV5 FOR TRAFFIC AND ROAD SIGN DETECTION

Sanyam Jain
*Dept. of Computer Science*
*Østfold University College*
Halden, Norway 1783
sanyamj@hiof.no

*Abstract*— This paper implements and investigates popular adversarial attacks on the YOLOv5 Object Detection algorithm. The paper explores the vulnerability of the YOLOv5 to adversarial attacks in the context of traffic and road sign detection. The paper investigates the impact of different types of attacks, including the Limited-memory Broyden-Fletcher-Goldfarb-Shanno (L-BFGS), the Fast Gradient Sign Method (FGSM) attack, the Carlini and Wagner (C&W) attack, the Basic Iterative Method (BIM) attack, the Projected Gradient Descent (PGD) attack, One Pixel Attack, and the Universal Adversarial Perturbations attack on the accuracy of YOLOv5 in detecting traffic and road signs. The results show that YOLOv5 is susceptible to these attacks, with misclassification rates increasing as the magnitude of the perturbations increases. We also explain the results using saliency maps. The findings of this paper have important implications for the safety and reliability of object detection algorithms used in traffic and transportation systems, highlighting the need for more robust and secure models to ensure their effectiveness in real-world applications.

## I. INTRODUCTION

Object detection is a critical task in computer vision that has applications in various fields, including autonomous driving, surveillance, and image search engines [1]. YOLOv5 (You Only Look Once version 5) is a popular deep learning-based one stage object detection algorithm that achieves state-of-the-art performance in terms of speed and accuracy [2]. However, recent studies have shown that deep learning models, including CNNs, are vulnerable to adversarial attacks, where an attacker can make small changes to the input image to mislead the model's predictions. Adversarial attacks on deep learning models pose a significant threat to the reliability and safety of object detection systems, especially in critical applications like traffic and transportation systems [3]. For example, an attacker could create adversarial traffic signs or road markings that could cause autonomous vehicles to misinterpret them, leading to accidents and potentially fatal consequences. In this paper, we investigate the vulnerability of the YOLOv5 object detection algorithm to adversarial attacks in the context of traffic and road sign detection. We evaluate the impact of various adversarial attacks on the accuracy of YOLOv5, including L-BFGS, FGSM, C&W, BIM, PGD, One Pixel Attack, and Universal Adversarial Perturbations attack [3, 4]. This paper aims to identify and analyze the effect of such attacks on predictions of the image nonresponding explanation on the image using GradCAM [5] which is a method to provide Explainability for the CNN model using layer-wise feature-saliency maps.

The remainder of the paper is organized as follows: Section II discusses the different types of adversarial attacks used in this paper and reproduction of results on these attacks on the used dataset. Section III defines dataset used and standard YOLOv5 architecture along with its training procedure. Section IV covers results and discussion based on the performed experiment. Finally, section V concludes with future directions.

## II. RELATED WORK

Each attack aims to perturb the input image in a specific way to cause misclassification by the model. The L-BFGS and C&W attacks are optimization-based and aim to find the smallest perturbation that causes misclassification, while FGSM, BIM, PGD, One Pixel Attack, and UAP attacks are gradient-based and iteratively update the image to maximize the loss function. The results show that all attacks were successful in causing misclassification of different techniques and applications. Following are key details about the attacks this paper uses and reference from [3,4]:

1. LBFGS is a gradient-based optimization algorithm that can be used to craft adversarial examples by minimizing the distance between the original image and the perturbed image subject to a constraint on the level of perturbation.
2. FGSM is a simple and computationally efficient adversarial attack that works by adding a small perturbation to each pixel of the original image in the direction of the gradient of the loss with respect to the image.
3. C&W attack is a state-of-the-art adversarial attack that uses an optimization-based approach to craft perturbations that result in misclassification of the target model with high confidence while minimizing the perceptibility of the perturbation.
4. BIM is an iterative attack that repeatedly applies the FGSM attack to the original image, but with a smaller step size, to generate a perturbed image that maximizes the loss of the target model.
5. PGD is a variant of the iterative FGSM attack that adds a projection step to ensure that the perturbed image remains within a specified distance from the original image in each norm.
6. One Pixel Attack is a simple but effective attack that works by modifying only one pixel of the original image to cause misclassification of the target model.
7. Universal Adversarial Perturbations are perturbations that can be added to any image and cause the target model to misclassify the image with high confidence. UAPs can be computed by averaging the gradients of the loss with respect to a set of images.

Saliency maps using Grad-CAM (Gradient-weighted Class Activation Mapping) visualization are techniques used to explain and interpret the decisions of deep neural networks [5]. Saliency maps highlight the most important regions of an input image that contribute to a model's prediction, while feature maps visualize the output of intermediate layers in the network, showing the activations of neurons and the representations learned by the model. Grad-CAM is a popular method for generating visual explanations, as it uses the gradients of the target class to highlight the important regions of the feature maps. These techniques provide insight into how the model makes predictions and can help identify potential biases or weaknesses in the model's performance. We use this technique to understand CAM maps before and after attack on the YOLOv5 model.

III. YOLOv5, Dataset and Training Methodology

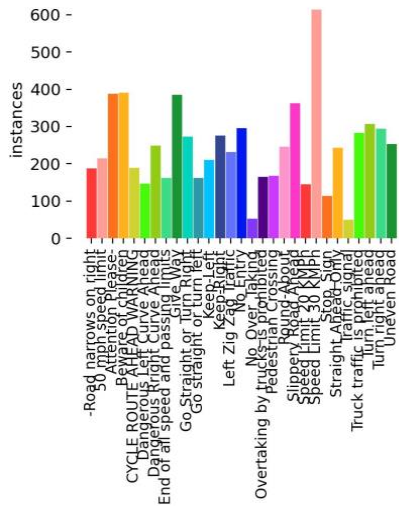

*Figure 1 Dataset distribution of classes*

YOLO (You Only Look Once) is a real-time object detection system that was introduced by Joseph Redmon and Ali Farhadi [6]. YOLOv5 [2,10] is an upgraded version of the previous YOLO versions and is designed to be faster and more accurate.

1. The backbone of YOLOv5 is the CSPDarknet53, which is a convolutional neural network (CNN) architecture designed for object detection tasks. CSPDarknet53 consists of 53 convolutional layers and is a modified version of the Darknet53 architecture. It is used to extract high-level features from the input image.
2. The bottleneck of YOLOv5 is the Spatial Pyramid Pooling (SPP) module. The SPP module is used to generate feature maps at multiple scales by pooling features from different regions of the feature maps. This allows YOLOv5 to detect objects of different sizes.
3. The head of YOLOv5 is composed of several convolutional layers and is responsible for generating the bounding boxes and class probabilities for the detected objects. The head takes the features generated by the backbone and bottleneck as input and outputs the final predictions.

Traffic and Road Signs Image Dataset is downloaded from [7]. It includes 29 classes for different traffic road signs (as shown in Figure 1). The dataset is available in YOLOv5 .TXT format as annotations and .JPG as image data. The dataset contains 10000 samples. Train-Val-Test ratio is 70:15:15. Images are stretched to 416x416 with augmentation applied while training. Further the training configuration is fixed as 300 epochs, 29 classes, CSP-DARKNET 53 as backbone, SGD Optimiser, weight decay 0.0005 and Leaky ReLU activation. In addition, the model is trained with Linux environment on AWS EC2 with p3dn instance family with 96 vCPU and memory of 76 GiB. The GPU used is NVIDIA V100 TensorCore. It is not the scope of this paper to dwell into great details of YOLOv5. For this paper, taking YOLOv5 as a deep learning model which is used to train and predict real images and perturbed images with respective adversarial attacks. Post training, inference take average of 10ms to predict and label the test images. The confusion matrix is shown in Figure 2. Once the model is trained and tested, weights are saved and loaded when required to perform the attacks, and visualizing class activation maps. "Yolov5-gradcam" [8] has been used to perform all required visualizations.

IV. Results and Discussion

The detailed results are shown in Table 1. Second column names all the attack names that are used while attacking the YOLOv5 model. Third column shows the prediction of the clean YOLOv5 trained model. Fourth column has predicted labels which matches the ground truth label before attack, and corresponding fifth column shows the predicted score. Now, sixth column has shown explanation before attack. Each image has three sub-images, where first one is the original image, second is the saliency map of the features that predicts the image to corresponding label, and final third sub-image has the most important region that supported model to predict the region coordinates. Seventh column shows attacked images that is equal to the sum of original image and noise which equals to perturbed image. Prediction score after attack shows the percentage of the predicted label (ninth column) which is not equal to ground truth or the clean model, which proves model is fooled with respective attacks. Moreover, this can be confirmed with the saliency maps on the same images but after attack which are called as explanation after attack. We release rest of the results and training curves for clean training in Appendix.

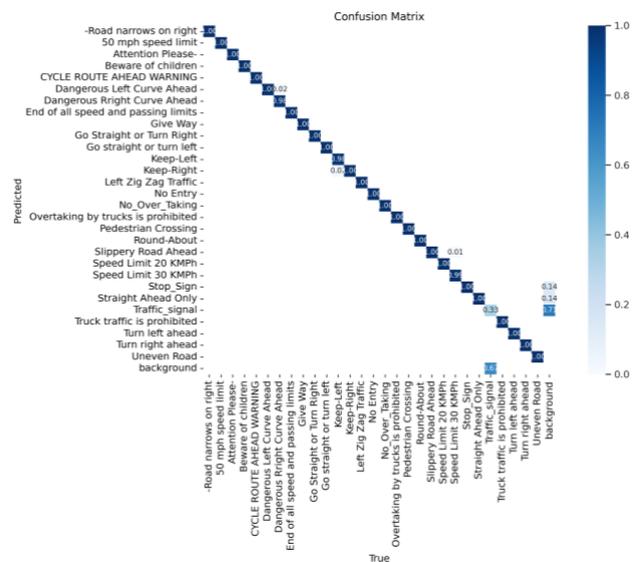

*Figure 2 Confusion Matrix of clean training YOLOv5*

## V. Conclusions

This short work proves that YOLOv5 is prone to adversarial attacks. Moreover, powerful attacks can even result wrong prediction with higher accuracy. One reasonable solution could be adversarial learning; however, it is reportedly proven that it is not the best way to mitigate adversaries. With the development of advanced one stage and two stage object detectors, there is a need of complete consideration about making such models fool-proof and robust with not only known attacks, but also known mitigation techniques. A future work could be finding robust adversarial learning mechanism that works specially for object detection models. Code and related python notebooks can be found here https://github.com/s4nyam/gradcam-yolo.

## Acknowledgment

Special thanks to Department of Computer Science, Østfold University College, Halden, Norway for providing enough resources to successfully complete this work.

## Appendix

Rest of the training results include F1 Curve, Labels Correlogram, P-Curve, R-Curve, PR-Curve, and miscellaneous results in Figure 3. Figure 4 comprises train and validation batch results.

| S.No. | Attack Name | YOLO prediction | Prediction Label before Attack | Prediction Score before Attack | Explanation Before Attack | YOLO pred After Attack | Prediction Score After Attack | Prediction Label After Attack | Explanation After Attack |
|---|---|---|---|---|---|---|---|---|---|
| 1 | L-BFGS | 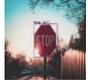 | Stop_Sign | 90% | 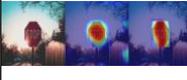 | 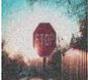 | NA | NA | 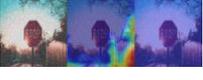 |
| 2 | FGSM | 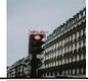 | Traffic_signal | 60% | 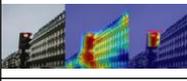 | 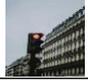 | NA | NA | 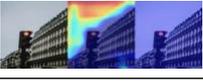 |
| 3 | C&W | 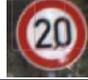 | Speed Limit 20 KMPh | 90% | 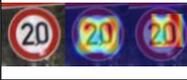 | 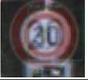 | 75% | 50 mph speed limit | 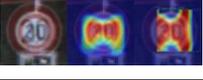 |
| 4 | BIM | 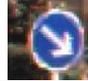 | Keep-Right | 83% | 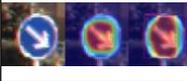 | 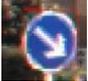 | 88% | Stop Sign | 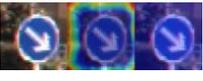 |
| 5 | PGD | 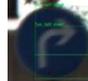 | Turn right ahead | 93% | 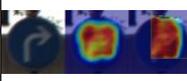 | 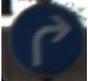 | NA | NA | 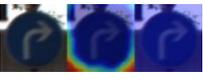 |
| 6 | OPA | 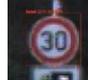 | Speed Limit 30 KMPh | 90% | 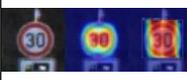 | 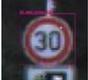 | 77% | 50 mph speed limit | 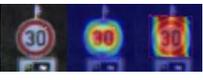 |
| 7 | UAP | 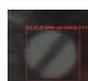 | End of all speed and passing limits | 88% | 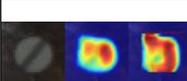 | 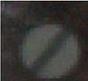 | NA | NA | 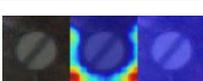 |

*Table 1. The table shows a detailed result chart of the adversarial attacks. "NA" resembles the place where the model was not able to recognize any of the 29 labels in respective input. The YOLOv5 prediction column shows the images which are predicted correctly and verified with prediction score before attack with corresponding label. Explanations before attack are produced using Grad CAM visualizations. The originally trained model is further fed with perturbed images to predict label and score. The model either is not able to predict anything or predicts wrong class with high accuracy (in column name YOLOv5 pred after attack). Further, explanation shows the model does not look over class specific features in such cases. And if it looks for features, those are the cases where it hallucinates over other region and put wrong label.*

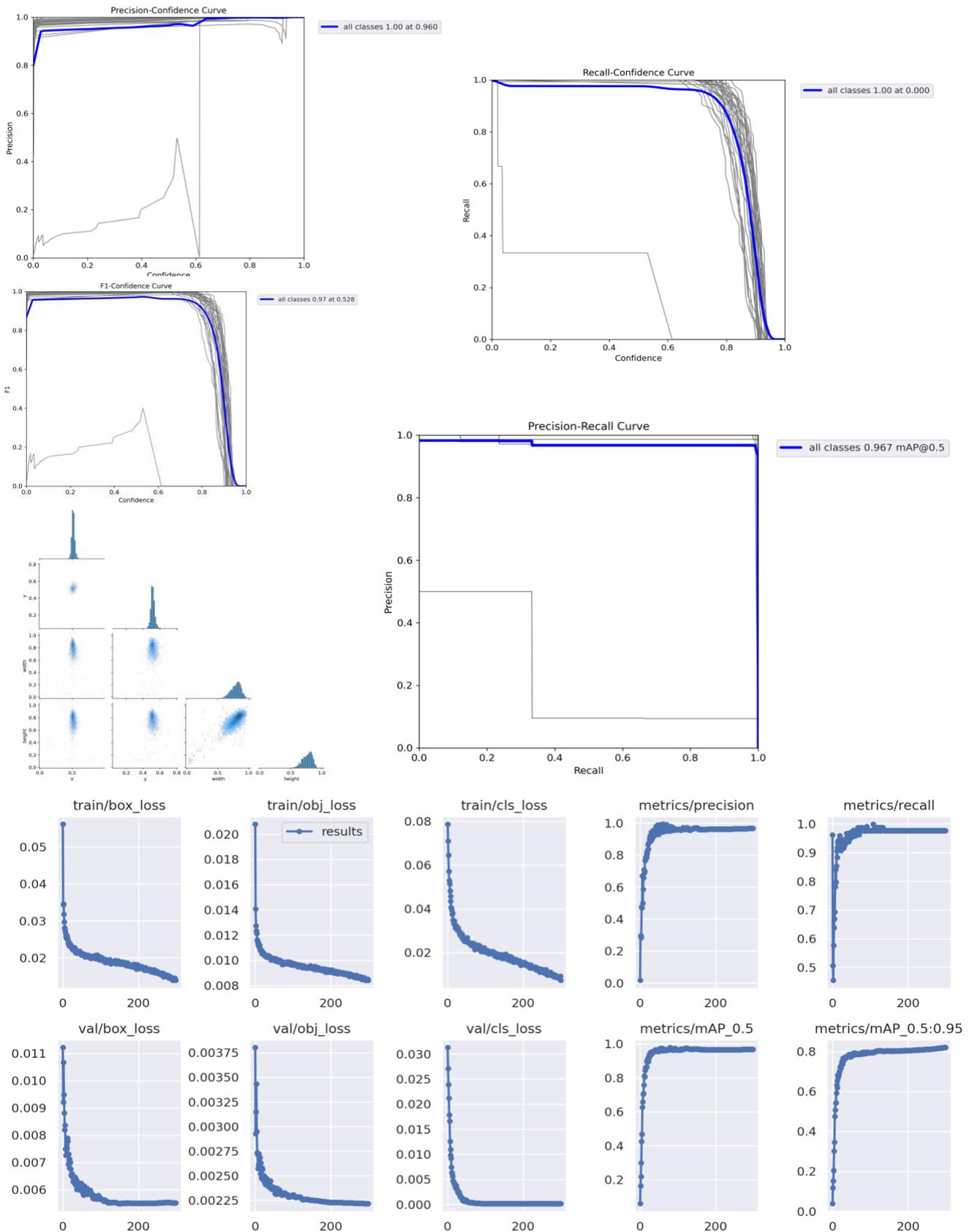

*Figure 3 Training results of clean version of YOLOv5 on the dataset. Results include P-Curve, R-Curve, PR-Curve, F1-Curve, Labels-Correlogram, Misc. Training Results*

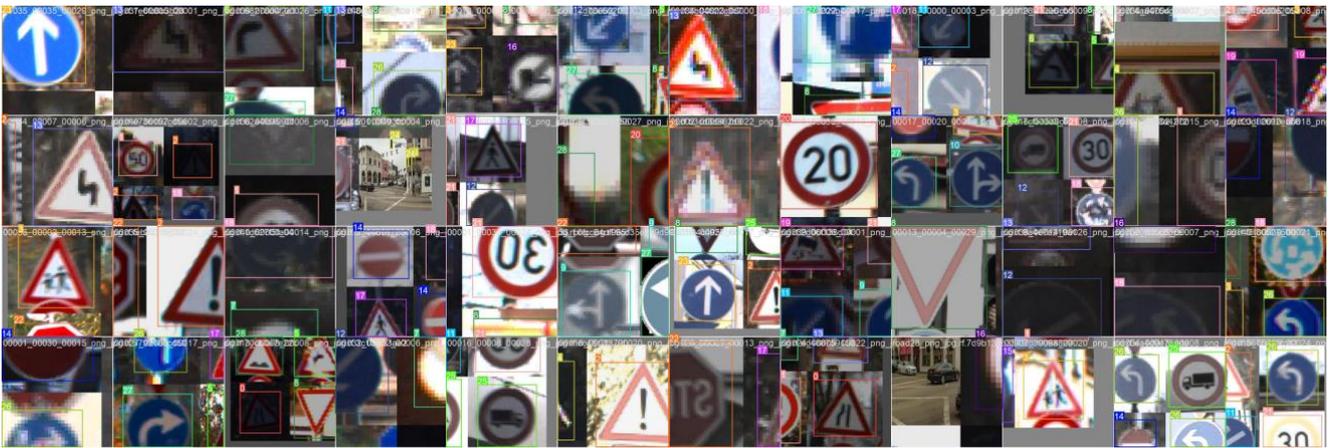

Train

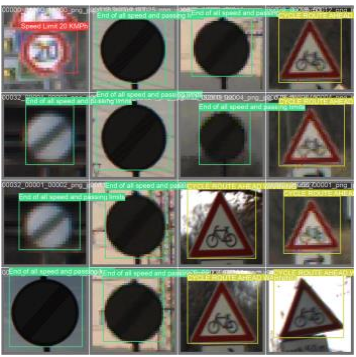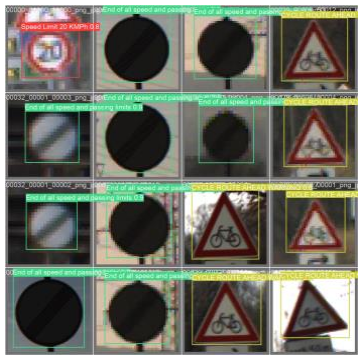

Validation Batch 0 Labels (Left)

Validation Batch 0 Prediction (Right)

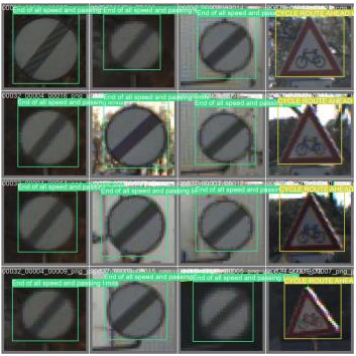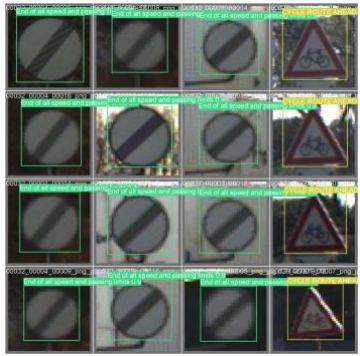

Validation Batch 1 Labels (Left)

Validation Batch 1 Prediction (Right)

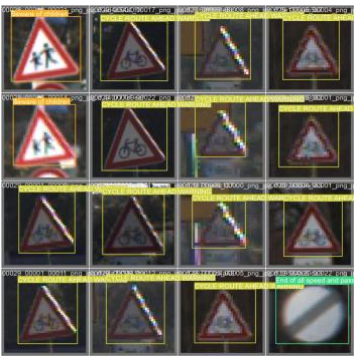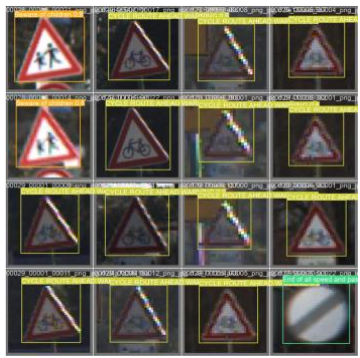

Validation Batch 2 Labels (Left)

Validation Batch 2 Prediction (Right)

*Figure 4 Train Batch and Validation Batch (with Labels and Predictions)*